\documentclass[11pt]{article}
\usepackage[utf8]{inputenc}
\usepackage[T1]{fontenc}
\usepackage{amsmath}
\usepackage[version=4]{mhchem}
\usepackage{stmaryrd}
\usepackage{graphicx}
\usepackage[export]{adjustbox}
\usepackage{lineno}
\graphicspath{ {./images/} }
\usepackage{latexsym}
\usepackage{float}
\usepackage{subfig}
\usepackage{relsize}
\usepackage{authblk}
\usepackage[hidelinks]{hyperref}
\usepackage{titling}
\predate{}
\postdate{}
\date{}

\title{Predicting Human Chess Moves: An AI Assisted Analysis of Chess Games Using Skill-group Specific n-gram Language Models }
\usepackage[a4paper, left=1in, right=1in, top=1in, bottom=1in]{geometry}

\usepackage[backend=biber, style=apa]{biblatex}
\author{Daren Zhong, Dingcheng Huang, Clayton Greenberg}
\begin{document}
\maketitle
\begin{abstract}
Chess, a deterministic game with perfect information, has long served as a benchmark for studying strategic decision-making and artificial intelligence. Traditional chess engines or tools for analysis primarily focus on calculating optimal moves, often neglecting the variability inherent in human chess playing, particularly across different skill levels.

To overcome this limitation, we propose a novel and computationally efficient move prediction framework that approaches chess move prediction as a behavioral analysis task. The framework employs n-gram language models to capture move patterns characteristic of specific player skill levels. By dividing players into seven distinct skill groups, from novice to expert, we trained separate models using data from the open-source chess platform Lichess. The framework dynamically selects the most suitable model for prediction tasks and generates player moves based on preceding sequences.

Evaluation on real-world game data demonstrates that the model selector module within the framework can classify skill levels with an accuracy of up to 31.7\% when utilizing early game information (16 half-moves). The move prediction framework also shows substantial accuracy improvements, with our Selector Assisted Accuracy being up to 39.1\% more accurate than our benchmark accuracy. The computational efficiency of the framework further enhances its suitability for real-time chess analysis.

\end{abstract}

\section{Introduction}
Chess, a deterministic game with perfect information, serves as a well-established benchmark for studying strategic decision-making and artificial intelligence. Its inherent complexity, with approximately $10^{120}$ possible game variations, makes it an ideal testbed for computational reasoning \cite{Shannon01031950}. Significant advancements in chess engines have evolved from Alan Turing’s pioneering 1948 Turochamp to modern systems such as Stockfish and AlphaZero \cite{urnSystematicAnalysis, rl2021}. These engines primarily focus on calculating optimal moves using deep search algorithms, reinforcement learning, and position evaluation functions, excelling in predicting the most advantageous continuations from given board states \cite{urnSystematicAnalysis, DBLP:journals/corr/abs-2008-04057}.

Most existing approaches to chess analysis focus on optimal move generation, which neglects the inherent variability and uncertainty characteristic of human decision-making \cite{Czech_Korus_Kersting_2021}. While determining optimal moves is useful to build the best chess engine, it does not adequately capture behavioral patterns. Conventional analysis often relies on evaluating human moves against computer-generated optimal moves, focusing on strategic correctness rather than the players' tendencies.

In addition, chess has been studied through the lens of natural language. Previous research has fine-tuned a GPT-2 model to learn chess using PGN notation and demonstrated that it can play and utilize high-level early-game strategies well \cite{DBLP:journals/corr/abs-2008-04057}. However, this research, too, focuses primarily on strategic correctness.

To address this gap, the proposed approach models chess move prediction as a language analysis problem rather than an optimal move calculation task. The framework utilizes n-gram language models, typically employed in natural language processing, to represent chess moves as sequences of words or tokens. Conceptualizing moves as linguistic elements enables the analysis of how move patterns differ across skill levels, revealing player tendencies that are not necessarily aligned with optimal play.

The primary objective is to develop a move prediction framework that reflects player behavior across a range of skill levels, from novices to experts. Players were categorized based on their ratings, and the model analyzed move sequences to identify their skill level. Rather than selecting the move with the highest confidence from all models, the framework applied the model most consistent with the player’s skill level, resulting in improved prediction accuracy and computational efficiency.

To validate the framework, we used data from the open-source chess platform Lichess. The dataset, containing games categorized by player rating, was preprocessed to eliminate redundant and inconsistent elements, ensuring reliable input for model training. The evaluation compared the framework's performance against a designed benchmark, using both Top-1 and Top-3 move prediction metrics. The results demonstrate that selectively applying skill-specific models improves accuracy, especially in scenarios where player moves exhibit inherent uncertainty.

The proposed framework offers a novel perspective on chess analysis, emphasizing play choices rather than solely focusing on optimal strategies. By identifying player skill levels and employing efficient move prediction techniques, the framework provides insights into how strategic choices evolve with increasing skill, offering a robust foundation for future research in human-centric chess analysis.

\section{Dataset and Data Processing}
\subsection{Dataset}
We utilized data from the open-source chess platform \href{http://lichess.org}{lichess.org}'s publicly available archive at \href{http://database.lichess.org}{database.lichess.org}, which contains all standard-rated games played monthly since January 2013. 

The original dataset was distributed in the PGN (Portable Game Notation) format and compressed using zStandard. PGN files encode each game as a combination of structured metadata (such as player ratings, time controls, and event information) and the full sequence of moves. The sequence of moves also contains elements such as move numbers, computer-generated evaluations (e.g., {[\%eval -0.29]}), symbolic annotations (e.g., "!" or "?"), and side variations. While these elements are useful for human analysis, they introduce inconsistencies that complicate language model training and are rarely needed in practical applications.

Games were placed in skill groups based on the average of each game's "WhiteRating" and "BlackRating" headers. We used the following bins:

\begin{center}
Level 1 (L1): <=1000\\ 
Level 2 (L2): 1000–1400\\
Level 3 (L3): 1400–1600\\
Level 4 (L4): 1600–1800\\
Level 5 (L5): 1800–2000\\
Level 6 (L6): 2000–2250\\
Level 7 (L7): >=2250    
\end{center}

This grouping scheme was informed by the empirical distribution of Lichess ratings and common distinctions in player skill levels. The rating distribution used for stratification is shown in Fig. 

\begin{figure}[h]
\centering
\includegraphics[width=1\linewidth]{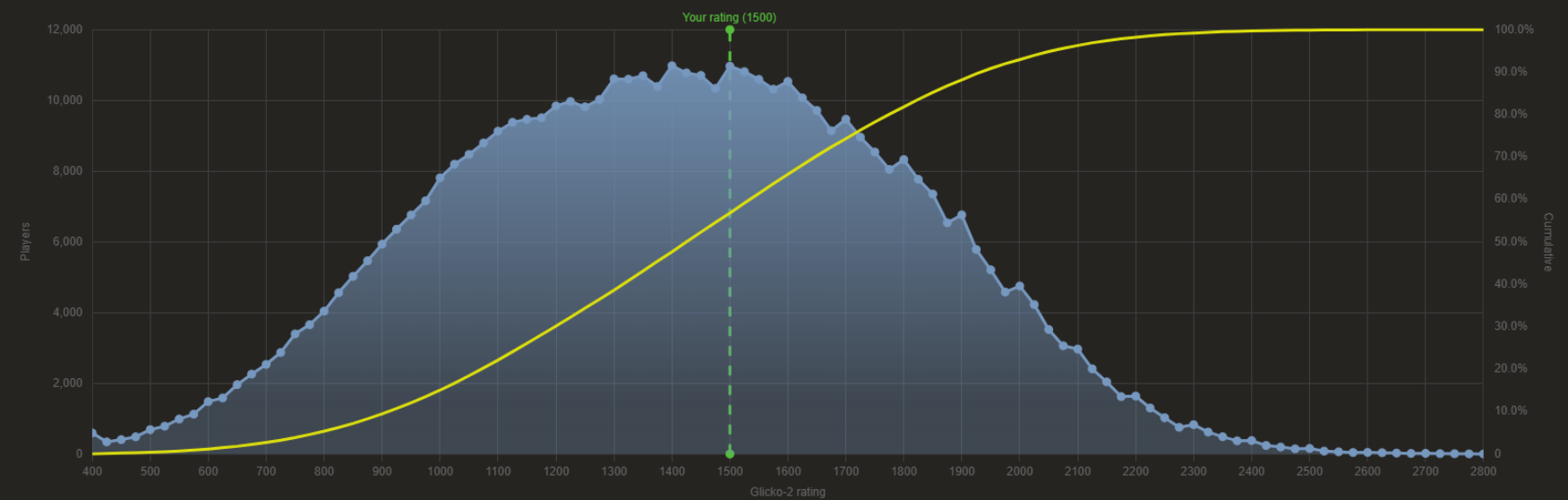}
\caption{The distribution considered for analysis}
\label{fig:2}
\end{figure}

To address hardware constraints when training language models, we took a random 10\% of each level's total games from July 2024 as training data. We also took 1000 games from August 2024 as testing data.

\subsection{Processing}
Pre-processing was a critical step in transforming raw PGN files into a form suitable for model training and inference. In particular, our goal was to standardize formatting to improve computational efficiency. Computational efficiency and an efficient input format improve the model's usability in real-world applications. Furthermore, because the models we used were n-gram, they are limited to a small number of words as history. Making the input format more efficient likely improved model accuracy as the model could access more relevant history.

After decompressing the data using a stream reader, we initially considered parsing the PGN files using the python-chess library. However, we ultimately opted to treat the PGNs as plain text, which significantly accelerated processing by bypassing legality checks and other parsing overhead.

\begin{table*}[h]
    \centering
    \begin{tabular}{|c|c|}
        \hline
        Metric & Purpose \\
        \hline
        Perplexity & Measures how well a model predicted a sequence;\\
        \hline
        Surprisal & Measures how well a model predicted a word;\\
        \hline
        16 Half-Move Accuracy & Assess performance of early game inference model selector\\
        \hline
        100 Half-Move Accuracy & Assess performance of full game inference model selector\\
        \hline
        Accuracy & A measure for how well the classifier worked\\
        \hline
        Average Error & Another measure for determining how well the classifier worked\\
        \hline
    \end{tabular}
    \caption{Metric and description of its purpose}
    \label{tab:1}
\end{table*}

We began by extracting and averaging the "WhiteRating" and "BlackRating" headers to assign each game to its corresponding skill group. Subsequently, we removed all metadata headers, side variations, and any elements not part of the main line of play. The games at this point still included move numbers and evaluations, as illustrated in the partially processed example below:\\

\textit{\small{1. e4 { [\%eval 0.2] } 1... e6 {[\%eval 0.13] } 2. Bc4 { [\%eval -0.31] } 2... d5 {[\%eval -0.28] } 3. exd5 {[\%eval -0.37] } 3... exd5 { [\%eval -0.31] } 4. Bb3 { [\%eval -0.33] } 4... Nf6 { [\%eval -0.35] } 5. d4 {[\%eval -0.34]} 5... Be7 { [\%eval 0.0] } 6. Nf3 { [\%eval 0.0] } 6... O-O { [\%eval -0.08] } 7. Bg5 { [\%eval -0.19] } 7... h6 { [\%eval -0.29] } 8. Bxf6 {[\%eval -0.36] } 8... Bxf6 { [\%eval -0.37] } 9. O-O { [\%eval -0.36] } 9... c6 {[\%eval -0.12] } 10. Re1 { [\%eval -0.17]} 10... Bf5 { [\%eval -0.04] }...}}\\

To produce a final standardized format, we removed all remaining non-essential elements, including move numbers, annotations, and computer evaluations. The end result was a clean, space-separated sequence of moves in algebraic notation:\\

\textit{\small{e4 e6 Bc4 d5 exd5 exd5 Bb3 Nf6 d4 Be7 Nf3 O-O Bg5 h6 Bxf6 Bxf6 O-O c6 Re1 Bf5...}}\\

Our pre-processing strategy ensured that each game was uniformly formatted by eliminating rare or irregular tokens that could negatively impact model training. It also produced a data format well-suited for real-world inference scenarios, where compact, standardized inputs are preferable for latency-sensitive applications.

Pre-processing yielded fourteen distinct corpora: seven training sets consisting of approximately 0.5 - 3 M games from July 2024, and seven corresponding test sets containing 1,000 games each from August 2024. Each corpus was organized according to the previously defined player rating levels.

\section{Move Prediction Framework}
\begin{figure*}[!t]
\begin{center}
\includegraphics[width=0.9\textwidth]{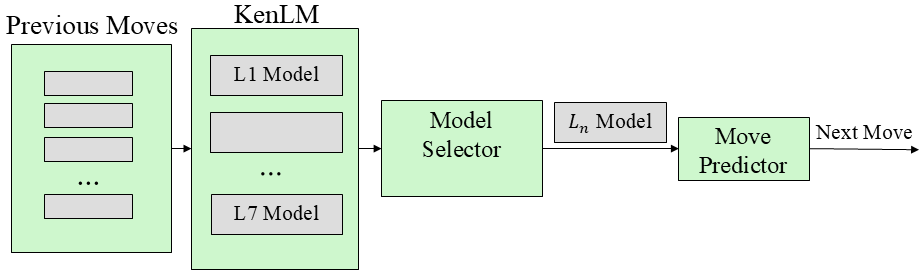}
\end{center}
\caption{Overview of the Chess Move Prediction Framework: The move prediction process relies on the model selected by the model selector, ensuring that the chosen model accurately reflects the player's skill level.
}
\label{fig: framework}
\end{figure*}

This section presents the move prediction framework, where the inference model processes the input chess game data and outputs probabilistic predictions, which are then fed into the Move Predictor to generate the next possible move. Fig. \ref{fig: framework} provides a visual illustration of the framework flowchart.

\subsection{Inference Model}
To perform inference on the processed chess game data, KenLM was adopted, a toolkit designed for efficient training and querying of n-gram language models. Our model selection was motivated primarily by its computational efficiency and low memory footprint, which made it particularly well suited to our available hardware resources \cite{kenlm}. Furthermore, KenLM has negligible hardware costs and latency for real-time users. 

We trained seven distinct n-gram language models, one for each of the rating-specific training sets from July 2024, using KenLM's lmplz function. Using KenLM’s query function, each model was then evaluated on all seven corresponding test sets from August 2024 \cite{queries}. This process produced 49 output files, each containing surprisal values for every move within each game, as well as overall perplexity scores for each test set.
\subsection{Inference Model Selector}
We developed a model selector that can classify each game based on the skill-level model that produced the lowest total surprisal. The total surprisal for each game for each language model was computed by aggregating the surprisal values assigned to each move. We were able to control the number of moves that were aggregated to classify a game. The model yielding the lowest cumulative surprisal was taken as the predicted skill level for that game.
\subsection{Move Predictor}
The selected model is used to predict the player's moves. Unlike approaches that rely on the global highest probability prediction across all models, our method exclusively utilizes the model that corresponds to the player's skill level. This approach is both computationally efficient and more accurate. Notably, the independence of each language model can result in global predictions that do not align with the player's proficiency. For example, when predicting moves for an L7 player, the global method might incorrectly select the L4 model, leading to inaccurate predictions.

\section{Experimental Setup}
A comprehensive evaluation of the framework was conducted using the validation set. This section details the hardware setup and the evaluation metrics used for assessment.
\subsection{Hardware Setup}
The models were trained using an AMD Ryzen 7 3700x with 32 GB of RAM. The Model Selector and Move Predictors ran on Windows Subsystem for Linux with 24 GB of RAM.
\subsection{Benchmark Setup}
We implemented the Top-1 move prediction benchmark by processing the input sequence of previous moves through all prediction models and selecting the move with the highest global confidence score among all language models. Similarly, the Top-3 move prediction benchmark identifies the three moves with the highest global confidence scores.
\subsection{Metrics}
Table \ref{tab:1} shows a table of the metrics used to measure performance. Note that half-moves are moves made by one player, for example, only "1. e4". This differs from a full move, which is "1. e4, e5". In the context of a language model, it is a single token.
\subsection{Experiment Design}
To evaluate the framework's performance, three experiments were conducted. First, the accuracy of the module selector was assessed to ensure accurate model selection corresponding to the player's skill level. Second, the move prediction accuracy was evaluated using the Top-1 Move Prediction, where the model predicts the opponent's next move based on the single highest-confidence output. Finally, we performed an evaluation using the Top-3 Move Prediction, which considers the three most likely moves generated by the model. This approach accounts for the inherent variability in player decisions, offering a more comprehensive assessment. By leveraging multiple high-confidence predictions, the Top-3 approach better captures the uncertainty present in real-world chess move prediction scenarios.

\section{Results and Discussion} 
\subsection{Inference Models}
\begin{figure}[h!]
\centering
\includegraphics[width=.41\textwidth, center]{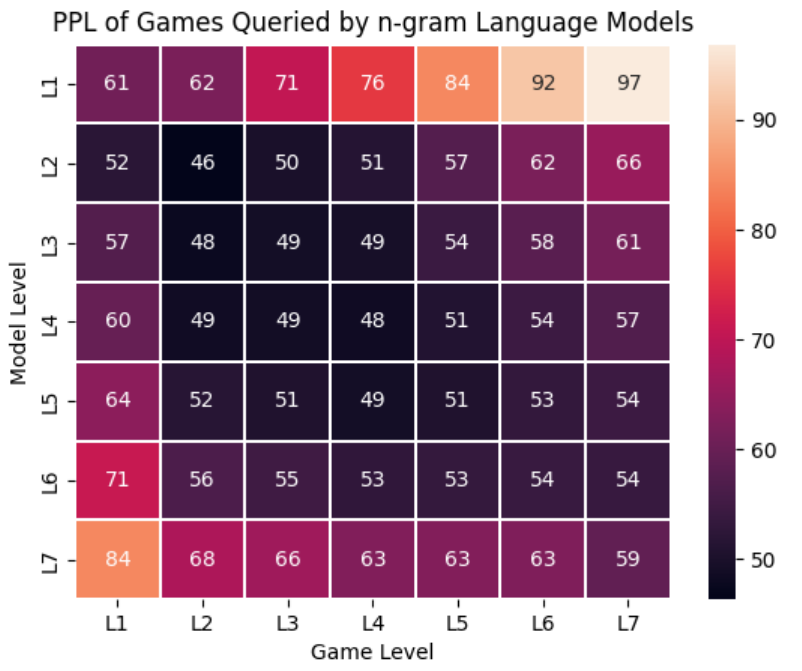}
\caption{Perplexity for each of the seven language models queried on each of the seven test sets}
\label{fig:PPL}
\end{figure}

We trained 7 5-gram language models on each of the 7 training sets. Each language model was its own level of play, aligning with the rating groups mentioned in Section 2.1. Fig. \ref{fig:PPL} presents a heatmap illustrating the perplexity values produced by each of the seven language models when evaluated on the seven corresponding test sets. In this visualization, the y-axis represents the model level, and the x-axis represents the game level. Ideally, the diagonals, where the model level matches the game level, should exhibit the lowest perplexities, as each model is expected to perform best on games from its corresponding rating level. The overall pattern in the heatmap supports this expectation, suggesting that the models are functioning as intended.

\begin{figure}[h]
\centering
\includegraphics[width=.5\textwidth, center]{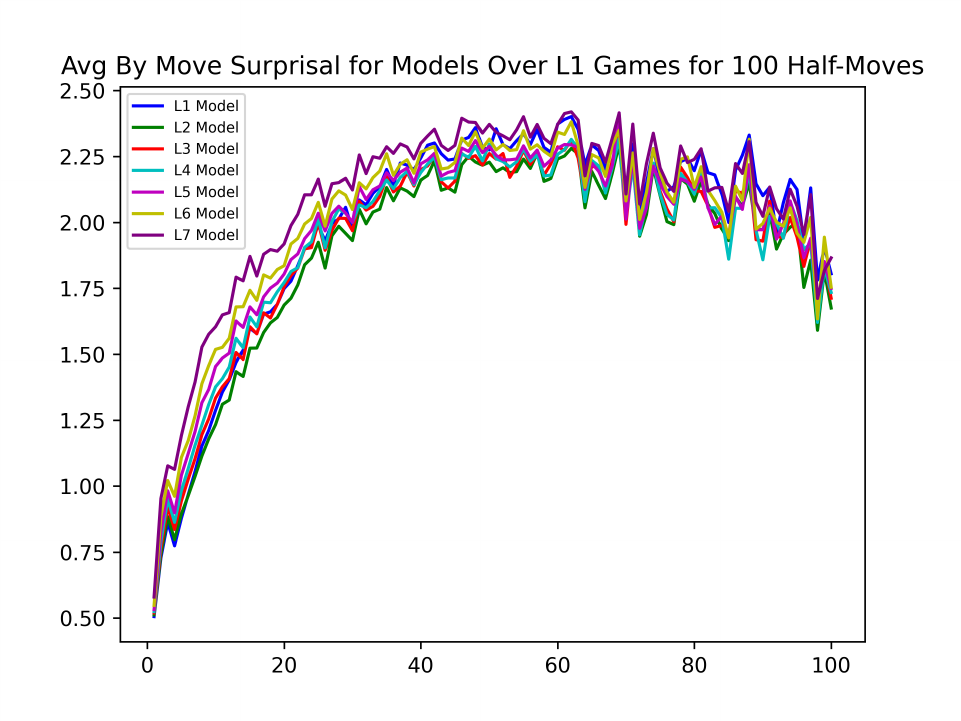}
\caption{Average by move surprisal for models over L1 games for 100 half moves}
\label{fig:avgMoveSurprisals}
\end{figure}

We also observed a decrease in the model’s certainty as games progressed toward the middle game. Fig. \ref{fig:avgMoveSurprisals} illustrates the average surprisal assigned to each move across L1 games. This reduction in certainty may stem from the increased number of viable move options typically available during the middle game.

\subsection{Inference Model Selector}

The model selector achieved an overall accuracy of 31.7\%. This was when evaluating the early game, or the first 16 half-moves. When the evaluation window was extended to include the first 100 half-moves, the overall accuracy decreased to 26.8\%. As shown in Table \ref{tab:2}, this decline is primarily due to a significant drop in classification accuracy for games at the extremes — L1 and L7 — when more of the game is considered. In contrast, games from rating levels L2 through L5 show improved classification accuracy when a larger portion of the game is used for inference.

\begin{table}[h]
    \centering
    \begin{tabular}{|c|c|c|}
        \hline
        Game Level & 16 Half Moves & 100 Half Moves \\
        \hline
        L1 & $37.2 \%$ & $22.3 \%$ \\
        \hline
        L2 & $35.4 \%$ & $39.0 \%$ \\
        \hline
        L3 & $23.3 \%$ & $23.9 \%$ \\
        \hline
        L4 & $24.5 \%$ & $27.9 \%$ \\
        \hline
        L5 & $26.6 \%$ & $31.8 \%$ \\
        \hline
        L6 & $32.0 \%$ &  $27.6\%$\\
        \hline
        L7 & $43.3\%$ & $15.3 \%$\\
        \hline
    \end{tabular}
    \caption{This table compares the accuracy the Inference Model Selector has for classifying each game into its correct game level using 16 half moves or 100 half moves}
    \label{tab:2}
\end{table}

We hypothesize that the sharp decrease in classification accuracy for L1 and L7 games when evaluating beyond the opening phase may be due to the distinct and unpredictable nature of player behavior in these groups. Novices (L1) often make erratic or near-random moves, especially as the game progresses, while advanced players (L7) tend to deviate from common patterns, employing highly sophisticated strategies that may not align with the statistical tendencies learned by the model. In both cases, the early game, which is typically more structured and studied, offers a more reliable basis for model inference.

\subsection{Top-1 Move Prediction}
\begin{figure}[!t]
    \centering
    \includegraphics[width=0.5\textwidth]{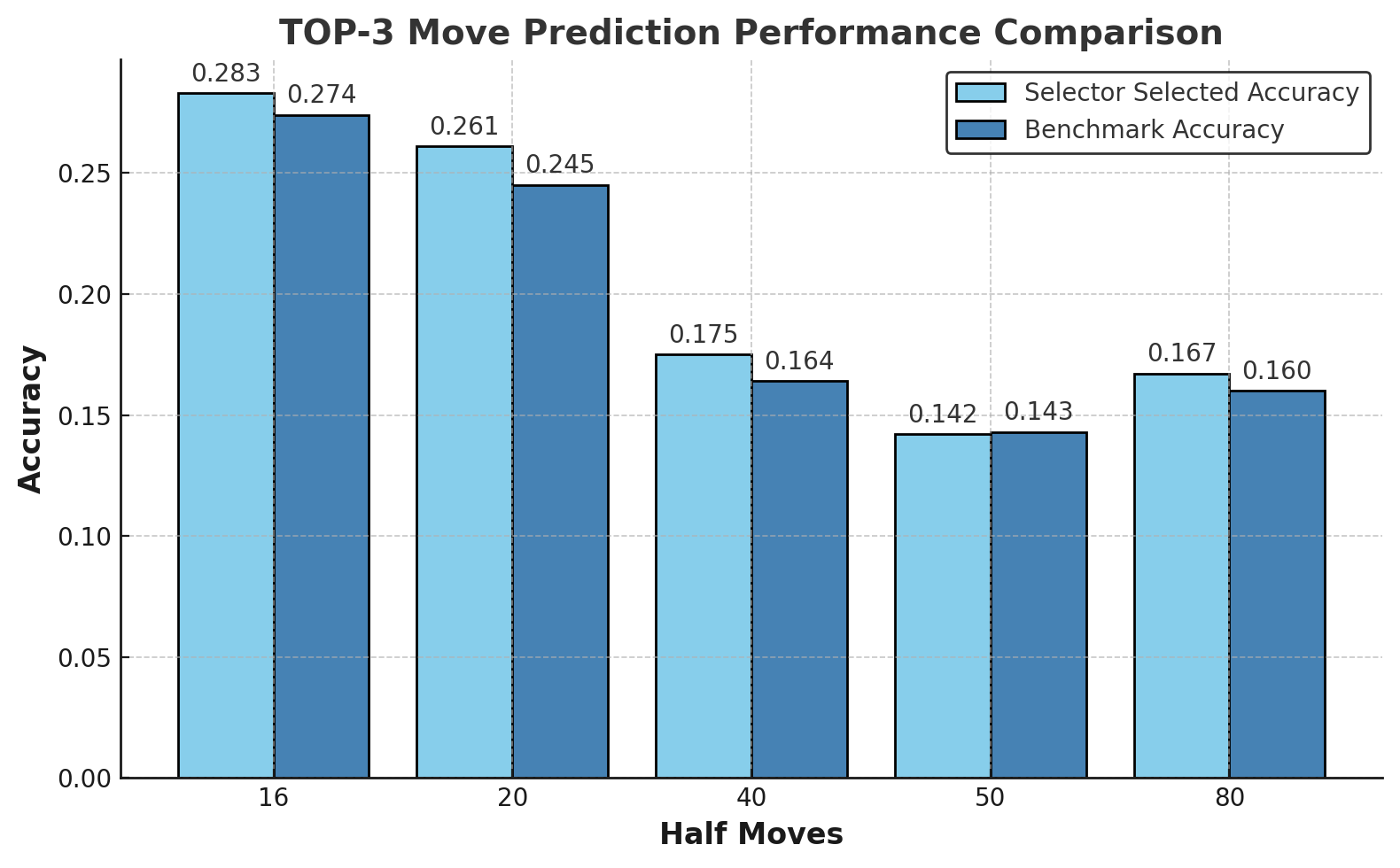}
    \caption{TOP-1 Move Prediction Performance Comparison. The plot shows the accuracy for different half moves, comparing the Selector Selected Accuracy and the Benchmark Accuracy.}
    \label{fig:top1_comparison}
\end{figure}

The evaluation results are presented in Fig. \ref{fig:top1_comparison}. The x-axis represents the number of previous half moves used for model selection. The results demonstrate that our framework consistently achieves higher move prediction accuracy compared to the benchmark, with improvements of up to 6.6\%.

One key observation is the relatively low overall accuracy, which reaches its minimum at 50 half moves. This trend suggests an increase in uncertainty during the middle game, likely due to the greater variability in strategic choices available to players at that stage.

Another important finding is that the accuracy improvement over the benchmark is generally modest. Specifically, at 50 half moves, the accuracy of our framework and the benchmark are nearly identical. To investigate this phenomenon, we conducted a deeper analysis and found that the games played by L1 and L7 players exhibit notably low prediction accuracy, significantly affecting the overall performance in the middle game.

This observation aligns with our earlier hypothesis that L1 and L7 players do not consistently follow model-identifiable patterns, often behaving as outliers. Consequently, the observed accuracy drop primarily stems from the inherent characteristics of players at these levels rather than limitations of the model itself.

\subsection{Top-3 Move Prediction}
The experiment is based on the premise that player move decisions involve inherent uncertainty. As a result, relying on Top-1 move prediction alone often leads to suboptimal accuracy. To better reflect the framework's predictive capability, Top-3 move prediction is utilized, as it accounts for multiple plausible moves rather than a single optimal choice.

\begin{figure}[!t]
    \centering
    \includegraphics[width=0.5\textwidth]{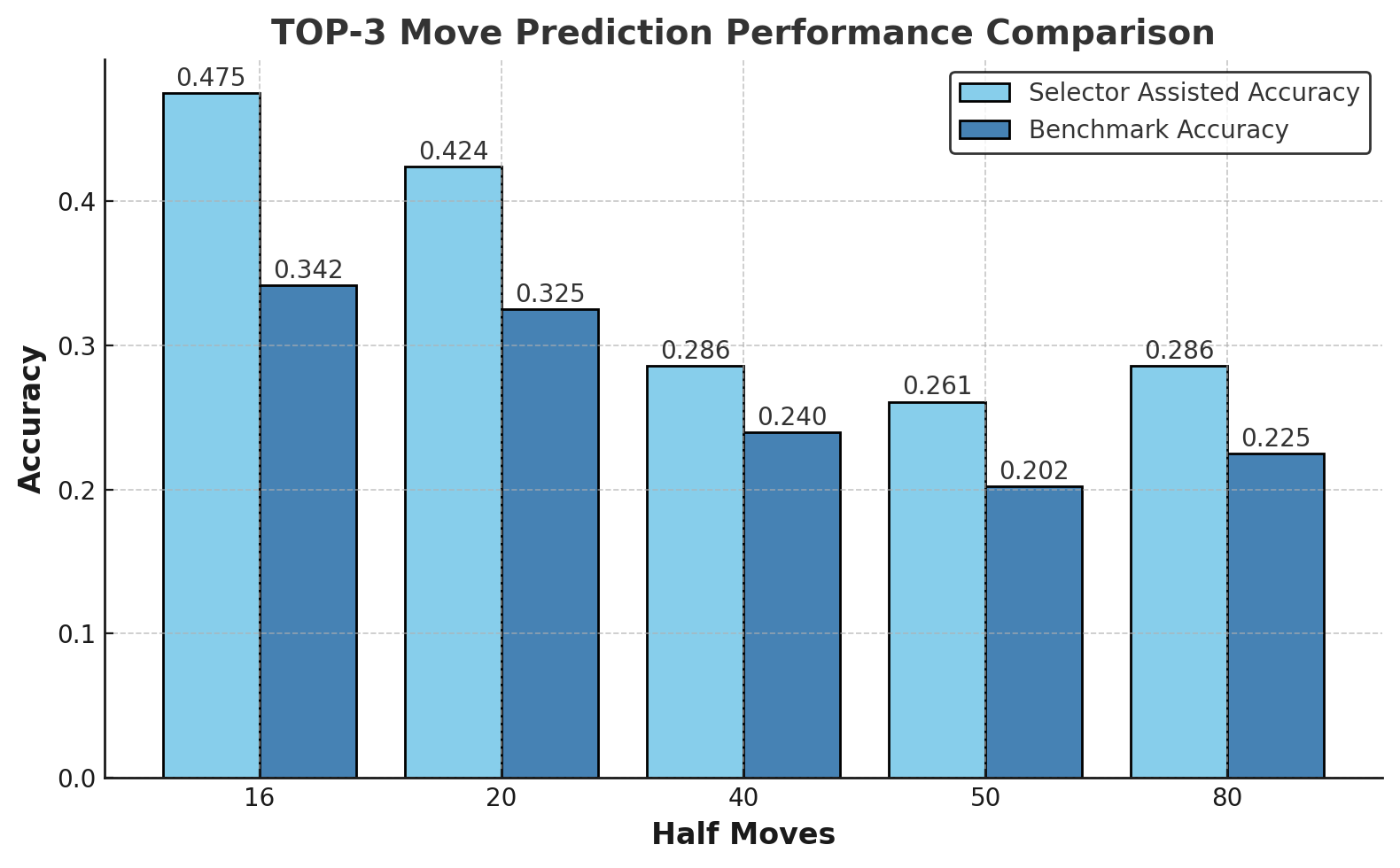}
    \caption{TOP-1 Move Prediction Performance Comparison. The plot shows the accuracy for different half moves, comparing the Selector Selected Accuracy and the Benchmark Accuracy.}
    \label{fig:top3_comparison}
\end{figure}

The prediction accuracy results are shown in Fig. \ref{fig:top3_comparison}. The framework demonstrates a substantial accuracy improvement over the benchmark, with an increase of up to 39.1\%. Categorizing the game into a specific player level based on previous move information significantly increases the accuracy of move prediction.

The results indicate that prediction accuracy reaches its minimum at 50 half moves, reflecting increased uncertainty in the middle game. The decline in accuracy is likely due to greater variability in strategic choices during this phase, making precise move prediction more challenging.

\section{Conclusion}
This paper introduced a novel and computationally efficient framework for predicting human chess moves, focusing on modeling player behavior rather than calculating optimal moves. Unlike traditional chess engines that primarily generate the best possible moves, the proposed framework uses n-gram language models to identify move patterns associated with specific player skill levels. By categorizing players into seven distinct skill groups, the framework dynamically adapts predictions based on the player’s demonstrated proficiency, offering a human-centric perspective on move prediction.

Evaluation on real-world game data from Lichess showed that the model selector module accurately classifies player skill levels with an accuracy of up to 31.7\% when utilizing early game information (16 half-moves). The move prediction framework also achieved accuracy improvements of up to 39.1\% compared to the benchmark. The ability to leverage skill-specific models for move prediction significantly enhances accuracy while maintaining computational efficiency, making the framework suitable for real-time chess analysis.

Analysis of performance across different game phases revealed a decrease in accuracy around the 50 half-move mark, indicating increased uncertainty during the middle game. The accuracy drop is most pronounced for novice (L1) and expert (L7) players, where move patterns deviate from intermediate-level play, reflecting greater variability in strategic decisions. The findings highlight the importance of accounting for behavioral diversity when designing predictive models in human-centric applications.

The proposed framework provides a structured approach to predicting chess moves by capturing skill-specific patterns, offering insights into human decision-making in competitive settings. By modeling how strategic choices evolve across skill levels, the framework sets the stage for future developments in move prediction and human behavior analysis in chess.

\section{Limitations}


One limitation of our research is that our inference models are unable to take into account more than 4 tokens of history. This might lead to failures to capture long-term context in player moves. 

Another limitation is that our models are oblivious to the game state, meaning they are unaware of the rules of chess, so the models could return illegal moves. Consequently, prediction performance is lowered. 

\section{Future Works}
Future work could explore capturing the sequential nature of moves to provide greater contextual understanding, potentially revealing insights into players’ overall strategies. This approach could enable the classification of openings or early-game strategies into distinct styles, such as aggressive or defensive. Such classifications may assist players across all skill levels in selecting openings that align with their preferred style of play. 

\clearpage
\printbibliography

@article{DBLP:journals/corr/abs-2008-04057,
  author       = {David Noever and
                  Matthew Ciolino and
                  Josh Kalin},
  title        = {The Chess Transformer: Mastering Play using Generative Language Models},
  journal      = {CoRR},
  volume       = {abs/2008.04057},
  year         = {2020},
  url          = {https://arxiv.org/abs/2008.04057},
  eprinttype    = {arXiv},
  eprint       = {2008.04057},
  bibsource    = {dblp computer science bibliography, https://dblp.org}
}

@misc{urnSystematicAnalysis,
	author = {Artturi Siven},
	title = {Systematic analysis of the evolution of chess computers},
	howpublished = {\url{https://urn.fi/URN:NBN:fi-fe2024083067265}},
	year = {2024},
	note = {[Accessed 18-05-2025]},
}

@inproceedings{kenlm,
    title = "Scalable Modified {K}neser-{N}ey Language Model Estimation",
    author = "Heafield, Kenneth  and
      Pouzyrevsky, Ivan  and
      Clark, Jonathan H.  and
      Koehn, Philipp",
    editor = "Schuetze, Hinrich  and
      Fung, Pascale  and
      Poesio, Massimo",
    booktitle = "Proceedings of the 51st Annual Meeting of the Association for Computational Linguistics (Volume 2: Short Papers)",
    month = aug,
    year = "2013",
    address = "Sofia, Bulgaria",
    publisher = "Association for Computational Linguistics",
    url = "https://aclanthology.org/P13-2121/",
    pages = "690--696"
}

@inproceedings{queries,
    title = "{K}en{LM}: Faster and Smaller Language Model Queries",
    author = "Heafield, Kenneth",
    editor = "Callison-Burch, Chris  and
      Koehn, Philipp  and
      Monz, Christof  and
      Zaidan, Omar F.",
    booktitle = "Proceedings of the Sixth Workshop on Statistical Machine Translation",
    month = jul,
    year = "2011",
    address = "Edinburgh, Scotland",
    publisher = "Association for Computational Linguistics",
    url = "https://aclanthology.org/W11-2123/",
    pages = "187--197"
}

@article{Czech_Korus_Kersting_2021, title={Improving AlphaZero Using Monte-Carlo Graph Search}, volume={31}, url={https://ojs.aaai.org/index.php/ICAPS/article/view/15952}, DOI={10.1609/icaps.v31i1.15952}, number={1}, journal={Proceedings of the International Conference on Automated Planning and Scheduling}, author={Czech, Johannes and Korus, Patrick and Kersting, Kristian}, year={2021}, month={5}, pages={103-111} }

@article{rl2021,
author = {David Silver  and Thomas Hubert  and Julian Schrittwieser  and Ioannis Antonoglou  and Matthew Lai  and Arthur Guez  and Marc Lanctot  and Laurent Sifre  and Dharshan Kumaran  and Thore Graepel  and Timothy Lillicrap  and Karen Simonyan  and Demis Hassabis },
title = {A general reinforcement learning algorithm that masters chess, shogi, and Go through self-play},
journal = {Science},
volume = {362},
number = {6419},
pages = {1140-1144},
year = {2018},
doi = {10.1126/science.aar6404},
URL = {https://www.science.org/doi/abs/10.1126/science.aar6404},
eprint = {https://www.science.org/doi/pdf/10.1126/science.aar6404},
}

@article{Shannon01031950,
author = {Claude E. Shannon},
title = {XXII. Programming a computer for playing chess},
journal = {The London, Edinburgh, and Dublin Philosophical Magazine and Journal of Science},
volume = {41},
number = {314},
pages = {256--275},
year = {1950},
publisher = {Taylor \& Francis},
doi = {10.1080/14786445008521796},


URL = { 
    
        https://doi.org/10.1080/14786445008521796
    
    

},
eprint = { 
    
        https://doi.org/10.1080/14786445008521796
    
    

}

}

\end{document}